%% file: root.tex
\documentclass[letterpaper, 10 pt, conference]{ieeeconf}  

\IEEEoverridecommandlockouts                              

\usepackage{amsthm,amsmath,amssymb}
\usepackage[ruled,linesnumbered]{algorithm2e}
\usepackage{algorithmic}
\usepackage{mathrsfs}
\usepackage{color}
\usepackage{cite}
\usepackage{multirow}
\usepackage{url}
\usepackage{xcolor}
\usepackage{xspace}
\usepackage{subfigure} 
\usepackage{graphicx}
\usepackage{comment}
\usepackage{threeparttable}

\usepackage[margin=0.75in]{geometry}

\usepackage{array}

\makeatletter
\let\NAT@parse\undefined
\makeatother
\usepackage[colorlinks,
            urlcolor=blue,
            linkcolor=blue,       
            anchorcolor=blue,  
            citecolor=green        
            ]{hyperref}

\newcommand{\pfilter}{{Voxel PFilter}\xspace}
\newcommand{\core}{OCC-VO\xspace}

\title{\LARGE \bf
\core: Dense Mapping via 3D Occupancy-Based Visual Odometry for Autonomous Driving 
}

\author{Heng Li, Yifan Duan, Xinran Zhang, Haiyi Liu, Jianmin Ji and Yanyong Zhang*
\thanks{The work is partially supported by the National Natural Science Foundation of China (No.62332016), Anhui Province Development and Reform Commission 2021 New Energy and Intelligent Connected Vehicle Innovation Project.}%
\thanks{* The corresponding author.}
\thanks{School of Computer Science and Technology, University of Science and Technology of China, Hefei, 230026, China
{\tt\small \{li\_heng, dyf0202, zxrr,  lhyakn\}@mail.ustc.edu.cn, \{jianmin, yanyongz\}@ustc.edu.cn}.}%
}%

\begin{document}

\maketitle
\thispagestyle{empty}
\pagestyle{empty}

\begin{abstract}



Visual Odometry (VO) plays a pivotal role in autonomous systems, with a principal challenge being the lack of depth information in camera images. This paper introduces \core, a novel framework that capitalizes on recent advances in deep learning to transform 2D camera images into 3D semantic occupancy, thereby circumventing the traditional need for concurrent estimation of ego poses and landmark locations. Within this framework, we utilize the TPV-Former to convert surround view cameras' images into 3D semantic occupancy. Addressing the challenges presented by this transformation, we have specifically tailored a pose estimation and mapping algorithm that incorporates Semantic Label Filter, Dynamic Object Filter, and finally, utilizes \pfilter for maintaining a consistent global semantic map. Evaluations on the Occ3D-nuScenes not only showcase a 20.6\% improvement in Success Ratio and a 29.6\% enhancement in trajectory accuracy against ORB-SLAM3, but also emphasize our ability to construct a comprehensive map. Our implementation is open-sourced and available at: \href{https://github.com/USTCLH/OCC-VO}{https://github.com/USTCLH/OCC-VO}.

\end{abstract}

\input{chapters/introduction}
\input{chapters/relatedworks}

\input{chapters/preliminaries}
\input{chapters/method}
\input{chapters/experiment}
\input{chapters/conclusion}

\bibliographystyle{IEEEtran}
\bibliography{IEEEabrv, references}

\end{document}

%% file: chapters/introduction.tex
\definecolor{barriercolor}{RGB}{255,136,132}
\definecolor{trafficconecolor}{RGB}{248,172,140}
\definecolor{drivablesurfacecolor}{RGB}{154,201,219}
\definecolor{sidewalkcolor}{RGB}{211,211,211}
\definecolor{terraincolor}{RGB}{153,102,51}
\definecolor{manmadecolor}{RGB}{128,128,128}

\section{INTRODUCTION} \label{INTRODUCTION}

Visual odometry (VO) and Visual Simultaneous Localization And Mapping (SLAM)  represent a fundamental technology in robotics and autonomous systems~\cite{chen2022overview}. Given that camera images do not have depth information, the fundamental task of VO and Visual SLAM is thus to concurrently estimate the ego poses and the landmark locations, with this process commonly referred to as Bundle Adjustment (BA)~\cite{triggs2000bundle}. 
BA is a complex optimization problem that can easily fail to achieve satisfactory results when faced with challenging circumstances, e.g., poor quality data, degenerate motions, and inadequate initial values for optimization~\cite{macario2022comprehensive}. Additionally, to meet the real-time computation requirement, the scale of the BA problem is typically controlled, with BA-based visual SLAM algorithms such as ~\cite{mur2015orb} that yield sparse maps without geometric details or semantic information.

Meanwhile, in the field of deep learning, significant progress has recently been made in the task of 3D semantic occupancy prediction~\cite{roldao20223d}. This task allows for the transformation of 2D image frames into 3D semantic occupancy, thus rectifying the shortcoming of lacking depth information in images.
By incorporating this task, we are able to simplify the core BA problem in traditional Visual SLAM, eliminating the need for simultaneous estimation of landmark locations. In other words, by treating the 3D semantic occupancy as a point cloud, the BA problem is thus transformed into a point cloud registration problem.

\begin{figure}
	\centering
    \includegraphics[width=0.85\linewidth]{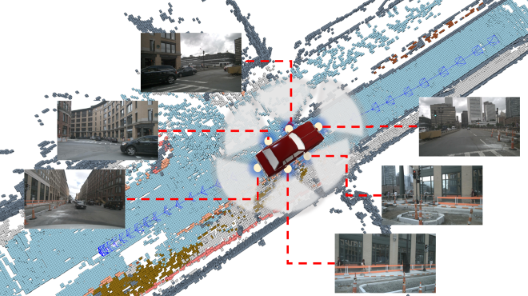}
    \begin{tabular}{*{16}{@{}l}}
    \raisebox{0.25ex}{\colorbox{barriercolor}{\rule{0pt}{0.1ex}\rule{0.1ex}{0pt}}} & \tiny \,\,Barrier \hspace{0.6em} &
    \raisebox{0.25ex}{\colorbox{trafficconecolor}{\rule{0pt}{0.1ex}\rule{0.1ex}{0pt}}} &  \tiny \,\,Traffic cone \hspace{0.6em} &
    \raisebox{0.25ex}{\colorbox{drivablesurfacecolor}{\rule{0pt}{0.1ex}\rule{0.1ex}{0pt}}} & \tiny \,\,Driveable surface \hspace{0.6em} &
    \raisebox{0.25ex}{\colorbox{sidewalkcolor}{\rule{0pt}{0.1ex}\rule{0.1ex}{0pt}}} & \tiny \,\,Sidewalk \hspace{0.6em} &
    \raisebox{0.25ex}{\colorbox{terraincolor}{\rule{0pt}{0.1ex}\rule{0.1ex}{0pt}}} & \tiny \,\,Terrain \hspace{0.6em} &
    \raisebox{0.25ex}{\colorbox{manmadecolor}{\rule{0pt}{0.1ex}\rule{0.1ex}{0pt}}} & \tiny \,\,Manmade \\
    \end{tabular}
    \vspace{-0.5em}
    \caption{Our approach transforms surround view cameras' image sequence into trajectories and global semantic maps. Such transformation can enhance scene understanding for downstream tasks in challenging environments.}
    \label{intro_fig}
    \vspace{-2.0em}
\end{figure}

However, the challenges arise because the 3D semantic occupancy differs from the original capture of the scene structure, e.g., Lidar scans. As a consequence, using such data to perform point cloud registration brings several issues.  Firstly, the coarse resolution of the 3D semantic occupancy introduces uncertainty in the estimation of landmark positions, subsequently affecting the accuracy of registration. Secondly, due to the imperfect neural network models, the landmarks may be inaccurately constructed or even partially missed. Lastly, it is important to differentiate between stationary environments and dynamic objects, especially in applications like autonomous driving, as their matches can result in less accurate pose estimation.  

In this work, we design and develop \core, a framework that takes surround view cameras' images as input and outputs a dense semantic map, which facilitates enhanced scene understanding for downstream tasks such as perception and navigation. Within this framework, we employ the open-source  3D semantic occupancy prediction module known as TPV-Former~\cite{huang2023tri}, to convert surround view camera images into 3D semantic occupancy.  

To cope with the issues in registration mentioned above, we devise a pose estimation and mapping algorithm tailored for 3D semantic occupancy. Specifically, we start with the well-known GICP algorithm~\cite{segal2009generalized} commonly used in Lidar-based SLAM as our baseline. This algorithm relies on feature matching and iterative optimization for the alignment of point clouds. Given the unique characteristics of the 3D semantic occupancy transformed from images, we introduce semantic constraints into our registration process, similar to ColorICP~\cite{park2017colored}. These semantic constraints prove to be highly effective in scenarios where geometric structures may be similar but with different semantics, such as road surfaces in autonomous driving contexts. Additionally, we implement a dynamic object filter to improve both map accuracy and pose estimation precision. Finally, during the mapping phase,  we leverage the idea in PFilter~\cite{duan2022pfilter} to eliminate unreliable points, building a more robust global semantic map. The end result is a fine-tuned pose estimation and a highly accurate map as depicted in Fig.~\ref{intro_fig}.

The evaluation of our approach is conducted using the Occ3D-nuScenes dataset~\cite{tian2023occ3d}, which is an extension of the nuScenes dataset~\cite{caesar2020nuscenes} augmented with voxel labels. We specifically focus on the training and validation sets captured by 6 surround view cameras in 2Hz. The diverse scenarios in the dataset, spanning different countries, lighting conditions, weather conditions, and environments, enable a thorough assessment of \core. Our method demonstrates superior performance in terms of accuracy and robustness in autonomous driving scenarios, compared to traditional Visual SLAM algorithms. In particular, when tested against ORB-SLAM3~\cite{campos2021orb}, our method demonstrates a Success Ratio that is improved by 20.6\% and shows a considerable gain in trajectory accuracy, reducing the absolute pose error by 29.6\%. Additionally, we assessed the completeness and precision of the map, proving its potential support for downstream tasks. However, we would like to point out that while our \core offers these advantages, it does require the presence of surround view cameras and incur more computation costs and longer latency. Thus in certain scenarios, it may not be a better alternative compared to other light-weight solutions. 

The main contributions of this paper are as follows:

\begin{itemize}

\item To the best of our knowledge, we design and develop \core which is the first to integrate 3D semantic occupancy with VO. This combination rectifies the shortcoming of lacking depth information
in images, enabling us to create dense and comprehensive maps, which facilitate enhanced scene understanding for downstream tasks in challenging environments.    
\item We have specifically designed a pose estimation and mapping module for the proposed framework, which addresses the inherent limitations of 3D semantic occupancy such as inference errors and uncertainty due to the coarse resolution. Under this framework, this allows us to achieve accurate and robust pose estimation as well as dense mapping.
\item Through trajectory evaluation, ablation study and map evaluation on Occ3d-nuScenes, our method demonstrates its superiority over traditional Visual SLAM algorithms, proving its robustness and accuracy in complex environments, even with low-frequency input. 
\end{itemize}





%% file: chapters/relatedworks.tex
\section{Related Work} \label{Related Work}

\subsection{3D Semantic Occupancy Prediction}

Recently, we have witnessed the rapid development  of 3D Object Detection based on Bird-Eye-View representations in autonomous driving~\cite{qian20223d}. However, relying solely on 3D object detection is insufficient for reconstructing realistic scenes. Unlike 3D Object Detection, 3D semantic occupancy prediction aims to reconstruct realistic scenes with rich semantic information. Incorporating semantic details offers more promising information for downstream tasks such as planning and SLAM.

Due to the similarities between the 3D semantic occupancy prediction task and the 3D Semantic Scene Completion (SSC) task, we will also incorporate the relevant SSC-related work into our research. Most of the related work relies on utilizing rich geometric information, such as Lidar point clouds~\cite{rist2021semantic, zhong2020semantic} and RGB-D images~\cite{song2017semantic, garbade2019two, li2020attention, wu2020scfusion}, which provides valuable depth and spatial information. In contrast, some recent works aim to reduce the reliance on geometric and depth information. MonoScene~\cite{cao2022monoscene} utilizes only a single RGB image to predict the semantic scene in indoor and outdoor environments. OCCDepth~\cite{miao2023occdepth} incorporates implicit depth cues from stereo images to aid in reconstructing 3D geometric structures. Furthermore, TPV-Former~\cite{huang2023tri} introduces a novel approach that leverages surround view cameras' images to construct tri-perspective representations, enabling the prediction of 3D semantic scenes.

\subsection{Visual SLAM} 


Visual SLAM is a branch of SLAM that primarily relies on visual information, such as images or videos, to estimate the ego pose and construct a map of the environment. It has gained significant attention due to the widespread availability of cameras in modern robotic platforms.

Traditional visual SLAM typically employ feature-based techniques. An example is ORB-SLAM~\cite{mur2015orb}, which detects and tracks features in images to estimate camera motion and reconstruct a sparse 3D map. These methods often rely on the extraction and matching of distinctive visual features across multiple frames. On the other hand, LSD-SLAM~\cite{engel2014lsd} is a direct method that estimates camera motion and reconstructs a semi-dense 3D map by directly minimizing the photometric error, exploiting all available pixel information.



In recent years, learning-based Visual SLAM has become extremely popular, such as CNN-SLAM~\cite{tateno2017cnn} DROID-SLAM~\cite{teed2021droid}. Besides, the impact of implicit 3D representation on visual SLAM has been significant, as demonstrated by the iMAP~\cite{zhu2022nice} and the NICE-SLAM~\cite{sucar2021imap}, which use RGB-D camera to SLAM. 

Recent years have also seen that Visual SLAM can benefit from occupancy prediction. By incorporating the 2D predicted semantic occupancy into the mapping and localization process, BEV-SLAM~\cite{ross2022bev} enhances the robustness and accuracy of the SLAM system. Our work, \core, takes this a step further by incorporating the predicted 3D semantic occupancy into the mapping and localization process. 



%% file: chapters/preliminaries.tex
\section{Preliminaries} \label{Preliminaries}

Before elaborating on the core components of our \core, it is crucial to outline some foundational algorithms. This section offers a brief overview of TPV-Former (Sec.~\ref{3D Semantic Occupancy Prediction}) and GICP (Sec.~\ref{GICP Algorithm}), both of which have a significant impact on the proposed  approach.


\subsection{3D Semantic Occupancy Prediction} \label{3D Semantic Occupancy Prediction}
We use TPV-Former~\cite{huang2023tri}  to obtain the 3D semantic occupancy. TPV-Former introduces a novel approach for characterizing 3D scenes utilizing a tri-perspective view (TPV) representation. 
To assemble TPV features, the process begins by feeding surround view cameras' images into an image backbone network for feature extraction. Following this, TPV-Former utilizes a cross-attention mechanism to assimilate image features into TPV features and a hybrid-attention mechanism to foster interactivity among three planes by sampling the corresponding positions in the three planes. Finally, dense voxel features are obtained by broadcasting each plane along the corresponding orthogonal direction and summing these features, and a lightweight decoder takes the voxel features to predict whether each voxel is occupied and the semantic label of the occupied voxel. 



\subsection{GICP Algorithm} \label{GICP Algorithm}
The GICP algorithm~\cite{segal2009generalized} is an enhanced version of the ICP algorithm~\cite{besl1992method}. This enhancement incorporates a Gaussian probability model into the ICP's cost function. The covariance matrix in this model performs a role analogous to weighting, which aids in mitigating the influence of outliers during the computation process. GICP retains the other segments of the algorithm unchanged, which assists in decreasing complexity and maintaining computational speed. 




In typical implementations, we obtain the point clouds matching results by utilizing a KD-tree to find the nearest neighbor points, denoted as $ A=\{a_i\}_{i=1,2,...,N} $ and $ B=\{b_i\}_{i=1,2,...,N} $, where $a_i$ and $b_i$ are corresponding points. At this point, we continue to assume that each point in both sets follows a Gaussian distribution, $a_i \sim \mathcal{N}(\hat{a_i}, C^{A}_i),\quad b_i \sim \mathcal{N}(\hat{b_i}, C^{B}_i)$.

Assuming that $\mathbf{T}$ is the transformation, we define the residual $ d^{(\mathbf{T})}_i = b_i - \mathbf{T} a_i $ and define $F_i(\mathbf{T}) = d^{(\mathbf{T})^T}_i (C_i^B + \mathbf{T} C_i^A \mathbf{T}^T)^{-1} d^{(\mathbf{T})}_i$. As the GICP algorithm proposes, we perform the following optimization procedure: 
\begin{equation}\small \label{cost function of GICP algorithm}
\mathbf{T^{\star}} = \mathop{\text{argmin}}_{\mathbf{T}} \sum_i F_i(\mathbf{T}),
\end{equation}
where $\mathbf{T^{\star}}$ represents the optimal transformation. 

%% file: chapters/method.tex
\section{Methodology} \label{Methodology}


\begin{figure}
	\centering
    \includegraphics[width=1.0\linewidth]{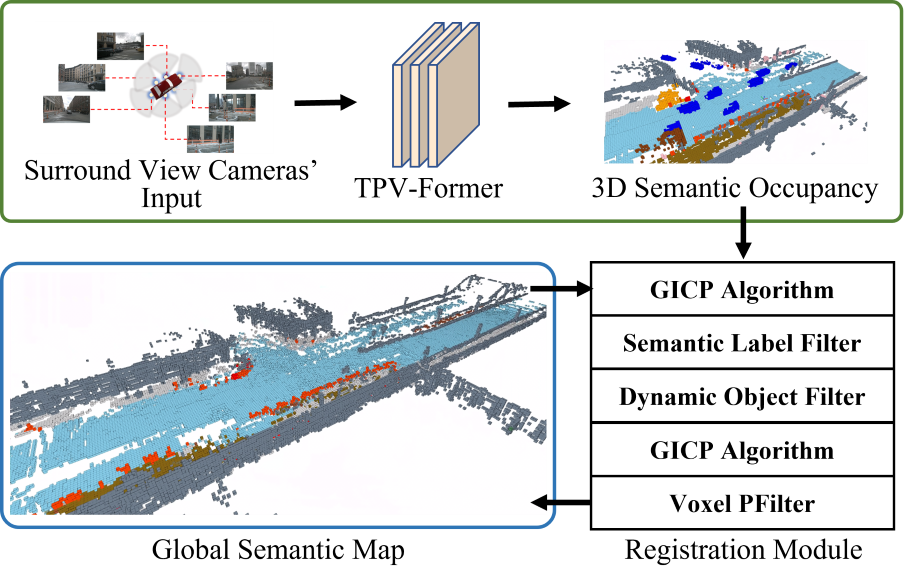}
    \vspace{-2.0em}
    \caption{Pipeline of our proposed \core.}
    \label{Pipiline}
    \vspace{-2.0em}
\end{figure}

\subsection{Overview} \label{Overview}
As shown in Fig.~\ref{Pipiline}, the 6 images captured by the surround view cameras at the same time are converted into 3D semantic occupancy
by TPV-Former mentioned in Sec.~\ref{3D Semantic Occupancy Prediction} at the beginning. The resulting 3D semantic occupancy is treated as point cloud, and the pose of each frame is estimated through registration between it and the global semantic map. In specific, we employed the GICP algorithm (Sec.~\ref{GICP Algorithm}) twice. At the beginning, the coarse correspondence between points is established. Following this, we engage the Semantic Label Filter (Sec.~\ref{Semantic Label Filter}) and Dynamic Object Filter (Sec.~\ref{Dynamic Object Filter}) to discard erroneous matches, thus refining the accuracy of the second GICP application. Once a precise pose is determined, we leverage the \pfilter (Sec.~\ref{Voxel_PFilter}) to merge the frame of data into the global semantic map, rectifying errors in TPV-Former's inference of the map for global consistency.


\subsection{Semantic Label Filter} \label{Semantic Label Filter}
In point cloud registration, a common cause of failure is the instability encountered on smooth planes, a common structure, e.g., the road,  in autonomous driving scenarios. Specifically, the alignment of point clouds tends to falter due to inadequate geometric constraints, causing a ``slip'' in the alignment~\cite{park2017colored}. This problem is hard to solve without introducing other sensors in point cloud-based SLAM systems~\cite{zhang2016degeneracy}. Fortunately, the point cloud from TPV-Former contains semantic labels besides geometric information.

Hence, we propose the use of a Semantic Label Filter by introducing semantic constraints to tackle this instability mentioned above. The Semantic Label Filter functions by eliminating pairs of points with mismatching semantic labels during the optimization process. This method effectively prevents erroneous point matches between various objects or surfaces from impacting the optimization solution.

To simplify the following formula, we continue our discussion by adopting Iverson bracket~\cite{iverson1962programming} defined as follows:
\begin{equation}\small
[P] = \left\{
\begin{array}{rcl}
1,     &      \text{if $P$ is True},\\
0,     &      \text{otherwise.}\\
\end{array} \right. 
\end{equation}


In this way, we employ $P_S(a_i, b_i)$ as the mathematical representation of the Semantic Label Filter. Within this function, $a_i$ is derived from the input 3D semantic occupancy, while $b_i$ is the corresponding point for $a_i$ in the global semantic map found through the first GICP. $P_S(a_i, b_i)$ is defined as $P_S(a_i, b_i) = (SL_{a_i} \text{ is equals to } SL_{b_i})$, where $SL$ represents the semantic label of $a_i$ and $b_i$.
Then we incorporate it into the cost function of GICP algorithm (Eq.~\ref{cost function of GICP algorithm}), specifically as follows: 
\begin{equation}\small
\label{cost function of Semantic Label Filter}
\mathbf{T^{\star}} = \mathop{\text{argmin}}_{\mathbf{T}} \sum_i F_i(\mathbf{T}) [P_S(a_i, b_i)].
\vspace{-1.0em}
\end{equation}



\begin{figure}[t]
	\centering
	\subfigure[Nearest points match] {\includegraphics[width = 0.2\linewidth]{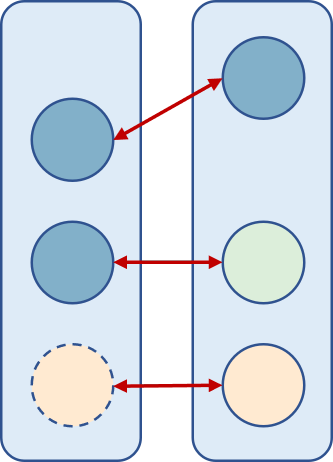}} \hfill
	\subfigure[Semantic Label Filter] {\includegraphics[width = 0.2\linewidth]{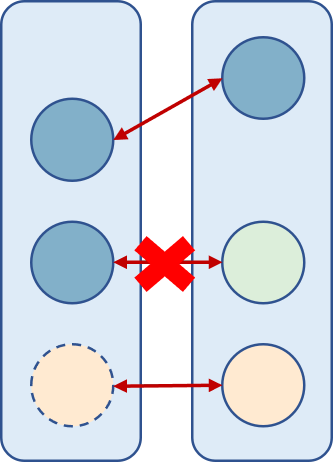}} \hfill
	\subfigure[Dynamic Object Filter] {\includegraphics[width = 0.2\linewidth]{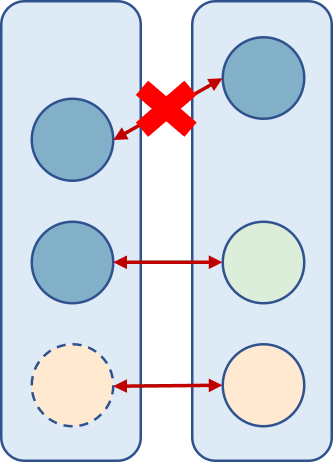}} \hfill
	\subfigure[\pfilter] {\includegraphics[width = 0.2\linewidth]{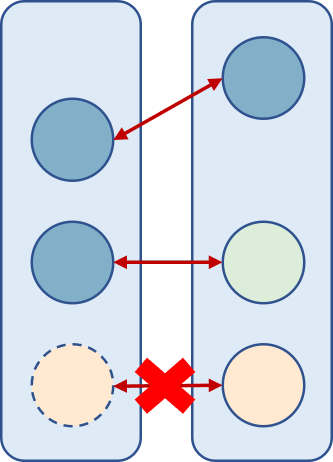}}
	\vspace{-0.5em}
    \caption{Three filters we propose. The rectangular boxes represent 3D semantic occupancy and the global semantic map separately, with each circle representing a specific point. The color of the circle represents  semantic labels, and the dashed circle border indicates that the point is transient with low p-Index defined in Sec.~\ref{Voxel_PFilter}. (a) shows three poor point-pair matches; (b) removes the match with different labels; (c) eliminates the one from dynamic objects; and (d) filters out the one containing a low p-Index value. 
    }
	\label{Filters}
	\vspace{-2.0em}
\end{figure}

\subsection{Dynamic Object Filter} \label{Dynamic Object Filter}

In autonomous driving scenarios, dynamic vehicles and pedestrians introduce significant disturbances to registration accuracy. One straightforward approach is using semantic labels to eliminate the occupancy of potential moving objects, such as various types of vehicles and humans. We refer to this method as Label-based Object Filter.


Yet, indiscriminately eliminating these potential dynamic objects can't always favorably contribute to pose estimation. Specifically, stationary objects often provide effective constraints for registration and  their removal may lead to scene degradation conversely. This issue becomes more severe when these objects are large vehicles such as engineering vehicles or buses, as they occupy a significant portion of the field of view of cameras.  

With the problem clearly outlined, we design the Dynamic Object Filter to enhance the performance of \core when dealing with highly dynamic scenarios. Specifically, objects with potential for motion are separated from both the 3D semantic occupancy and the global semantic map using semantic labels. Each object is then subjected to point cloud clustering, with these clusters subsequently treated as unified objects. Utilizing the transformation results from the first GICP shown in Fig.~\ref{Pipiline}, we compare the position of each object. This allows us to identify relative displacements and decide if an object is dynamic and should be removed. Employing the static part of each input, a refined registration is conducted, leading to enhanced precision in pose estimation.

Using the aforementioned algorithm, we have acquired the point set $DA$ corresponding to dynamic objects in the 3D semantic occupancy, as well as the point set $DB$ associated with dynamic objects in the global semantic map. Thus the Dynamic Object Filter can be defined as $P_D(a_i, b_i) = (a_i \notin DA \text{ and } b_i \notin DB)$. Similar to Sec.~\ref{Semantic Label Filter}, the cost function, i.e., Eq.~\ref{cost function of Semantic Label Filter}, can be extended as:  
\begin{equation}
\small
\label{cost function of Dynamic Object Filter}
\mathbf{T^{\star}} = \mathop{\text{argmin}}_{\mathbf{T}} \sum_i F_i(\mathbf{T}) [P_S(a_i, b_i)][P_D(a_i, b_i)].
\end{equation}


\subsection{\pfilter } \label{Voxel_PFilter}

Considering that for the same object or surface, the 3D semantic occupancy predicted between adjacent frames might have inconsistent grid representations, which is quite common in this field of work. As a result, we suggest the incorporation of \pfilter in the registration process to merge the more reliable points in the 3D semantic occupancy into the global semantic map. This modification aims to 
maintain the global consistency of the map and correct noise induced by network inference.

\pfilter is an improved version of PFilter~\cite{duan2022pfilter}, a feature selection algorithm for Lidar-based SLAM. To adapt to the problems of this work, we redefine the metric, i.e., p-Index, proposed in PFilter. In our implementation, the p-Index is a metric designed to determine the persistence level of a voxel.
For clarify, the meaning of the voxel being occupied in this section is that there is a point in the voxel, which is the data format we actually deal with in \core. 
During the mapping process, two attributes are recorded for each occupied voxel: the time when the voxel is first predicted as occupied, denoted as $t_0$, and the number of times it has been marked as occupied, denoted as $f$. The p-Index of voxel $v$ at time $t$ is simply defined as $\frac{f}{t - t_0}$. 

The voxel with a high p-Index value is consistently predicted to be occupied, indicating that it has a high probability of actually being occupied. This is what we call the persistent voxel. In contrast, the transient voxel, often caused by the network's erroneous prediction, exhibits lower values. 


Similar to Sec.~\ref{Dynamic Object Filter}, we define \pfilter as 
$P_V(b_i) = (\text{p-Index}(v)>0.5|b_i\text{ in }v)$, where 0.5 is derived from experience. Now substituting \pfilter into Eq.~\ref{cost function of Dynamic Object Filter} yields: 
\begin{equation}\small
\label{cost funtion of PFilter}
\mathbf{T^{\star}} = \mathop{\text{argmin}}_{\mathbf{T}} \sum_i F_i(\mathbf{T}) [P_S(a_i, b_i)][P_D(a_i, b_i)][P_V(b_i)].
\end{equation}

Furthermore, a downsampling procedure is applied to the persistent points. This involves conducting a weighted average of three-dimensional coordinates within each grid based on the p-Index. Only the points newly generated through this downsampling are retained, with a fresh p-Index calculation performed for them. 


%% file: chapters/experiment.tex
\section{Experiments} \label{Experiments}

\input{table/trajectory_evaluation}
\input{table/ablation_study}

We first introduce the setup of our experiment in Sec.~\ref{Experimental Setup}. Then we evaluate \core using the Occ3D-nuScenes datasets, contrasting its performance with the traditional method, i.e., ORB-SLAM3~\cite{campos2021orb} and learning-based method, i.e., DROID-SLAM~\cite{teed2021droid} in Sec.~\ref{Trajectory Evaluation}. Ablation studies are performed to underscore the potency of our methodology in Sec.~\ref{Ablation Study}. Further, we illustrate that \core can adeptly construct a comprehensive and accurate 3D semantic map in autonomous driving scenarios in Sec.~\ref{Map Evaluation}. Lastly, we conduct a execution time analysis, demonstrating the real-time capabilities of our algorithm in Sec.~\ref{Execution Time Analysis}.

\subsection{Experimental Setup}  \label{Experimental Setup}

In our experiments, we evaluate \core using the validation sets of the Occ3D-nuScenes dataset~\cite{tian2023occ3d}.
The Occ3D-nuScenes dataset serves as a comprehensive benchmark for 3D semantic occupancy prediction. Building upon the nuScenes dataset, it incorporates voxel labels and is designed specifically for autonomous driving applications. The dataset offers a varied and realistic collection of sensor data from multiple urban settings, positioning it as an ideal benchmark to gauge the performance and robustness of our algorithm across diverse conditions. Distinctive in its breadth and detail, the dataset covers over 1000 scenes, each lasting about 20 seconds, which spread across different countries, lighting settings, weather variations, and environments.

On the training sets, we train the TPV-Former on the 1600x900 image sequences captured at 2Hz by six surround view cameras. These sequences, spanning a total of 700 scenes, are annotated with voxel labels serving as the ground truth. Referring to TPV-Former's implementation, we set the occupancy prediction range as [$-40m, 40m$] for the X and Y axes, and [$-1.0m, 5.4m$] for the Z axis. The final output 3D semantic occupancy is presented in a 200x200x16 shape with a voxel size of 0.4m.

In the subsequent sections, we focus on experiments conducted on 150 validation sequences. All experiments employ 2Hz image sequences, as provided by Occ3D-nuScenes, with a Dynamic Object Filter displacement threshold set at 2 meters. In addition, ORB-SLAM3 operates at 12Hz because it can't work with 2Hz input.

\subsection{Trajectory Evaluation}  \label{Trajectory Evaluation}

We employ the Root Mean Square Error (RMSE) of Absolute Pose Error (APE)~\cite{grupp2017evo} of the predicted trajectories as the primary evaluation metric. Furthermore, due to the complexity of autonomous driving scenarios, these algorithms might yield results that deviate significantly from the correct trajectory in some cases, such as numerous dynamic objects and poor light conditions. Considering that the average APE for most samples is around 0.2 meters, any instance with an APE greater than 5 meters can be deemed a failure. Thus, we have introduced a success ratio metric, which represents the proportion of results with an APE of less than 5 meters. This is employed to filter out the inferior results and subsequently compute the aforementioned RMSE. 


Table~\ref{trajectory_evaluation_table} present our experimental results. It's evident that \core achieves a higher success ratio and a reduced APE. Specifically, \core boasts a success ratio of 99.3\% and an RMSE of APE at 0.140 meters. This represents a 20.6\% increase in success ratio and a 29.6\% improvement in trajectory accuracy compared to ORB-SLAM3, even when fed with a lower frequency input. Based on our analysis of experimental samples, the boost in success ratio is primarily observed in scenarios like rapid turns and poor lighting conditions, while the decrease in APE is attributed to robustness in complex environment such as numerous dynamic objects and extensive occlusions. In addition, we use the 3D semantic occupancy ground truth provided by Occ3D-nuScenes as the input of registration module, getting a success ratio of 100\% and an RMSE of APE at 0.122 meters, which shows the potential of \core, especially when applied with more accurate predict networks. 

\subsection{Ablation Study} \label{Ablation Study}



We next present the results of the ablation study. As shown in Table~\ref{ablation_study_table}, the experiment with different filter combinations exhibits varying levels of success ratio and accuracy. When no filters were employed, we observed the baseline performance with a success ratio of 97.3\% and an RMSE of APE at 0.220 meters. As we incrementally introduced the three filters, there was a consistent uptrend in the algorithm's performance: the success ratio advanced from 97.3\% to 98.0\% and finally to 99.3\%, while the RMSE of APE progressed from 0.220 meters to 0.206 meters, then to 0.173 meters and settled at 0.140 meters with the full filter set. Notably, when comparing the Dynamic Object Filter with the Label-based Object Filter, the latter did not enhance the success ratio and even resulted in an RMSE of APE decrease to 0.218 meters, underscoring the superior efficacy of the Dynamic Object Filter's design.


\subsection{Map Evaluation} \label{Map Evaluation}

\input{table/map_compare}
\input{table/map_evaluation}

To highlight the powerful capabilities of \core in dense outdoor mapping, we perform map evaluations. 
In this experiment, we generate map ground truth using the 3D semantic occupancy ground truth and the pose ground truth. The algorithm's output is assessed using accuracy, accuracy ratio and completion ratio. Accuracy quantifies the RMSE of the distance between sampled points from the reconstructed map and the nearest map ground truth point. Precision and completion ratio separately measure the proportion of points in the output reconstructed properly and the proportion of points in the map ground truth that are reconstructed. In our calculation, a map point is deemed ``reconstructed'' if its distance to the nearest reconstructed point is less than 0.4 meters, since the voxel size is 0.4 meters. 

In our experiment, we find that even outdoor-capable visual SLAM algorithms, like DROID-SLAM~\cite{teed2021droid}, fail to produce accurate and complete maps in such autonomous driving scenes. Thus we only conducted experiments under the conditions of \core with and without \pfilter.
As shown in Table~\ref{map_evaluation_table}, when using \pfilter, the algorithm exhibited higher accuracy and precision, but due to the filtering of transient points, completeness slightly decreased, with respective values of 0.111 meters, 72.4\% and 72.5\%. Without \pfilter, the algorithm get an 87.5\% completeness, but decrease in accuracy and precision, with values of 0.125 meters and 57.2\% respectively. The results show that the \core maintains its ability to perform accurate and dense mapping in complex autonomous driving scenarios, thereby demonstrating the capability to assist with downstream tasks such as navigation. As illustrated in Table~\ref{map_compare_table}, we present the qualitative visual results of our algorithm. These visual representations show that many static semantic details, such as vegetation, driveable roads, and barrier, are accurately reconstructed by our algorithm. 



\subsection{Execution Time Analysis}\label{Execution Time Analysis}

Owing to the high video memory demands of the 3D semantic occupancy prediction network, we conducted training and testing on a server equipped with 4 Intel Xeon Gold 6230R CPUs @ 2.10GHz, 8 NVIDIA A100 GPUs and 754 GB RAM. 
For the registration module, testing was executed directly on a Intel i9-13900K CPU @ 3.00GHz and 128 GB RAM personal computer. Given that the 3D semantic occupancy prediction network primarily relies on GPU performance, while the registration module leans more towards CPU capabilities, it is reasonable to evaluate and analyze their execution time on separate hardware platforms.

A detailed execution time analysis is shown in Table~\ref{execution_time_analysis_table}. The 3D semantic occupancy prediction network consumes 5401MB of GPU memory and requires 267 ms per inference. The registration module takes another 371 ms for each computation, pointing to possible limitations of the current Python-based implementation. Notably, both modules can operate in a pipelined manner. Thus, even with a 2Hz input, they ensure seamless processing without causing a backlog in the queue. Within the scope of our study, our attention has been on the integration and assessment of these modules in the \core system, highlighting that both the trajectory accuracy and map completeness are commendable. As future endeavors, for faster execution speed, we can refining the network's efficiency and migrating the registration to C++. 

\input{table/runtime_analysis}


%% file: table/trajectory_evaluation.tex
\begin{table*}[]
\centering
\begin{threeparttable}
\caption{Success Ratio and RMSE of APE [m] of various methods.}
\vspace{-1.0em}
\label{trajectory_evaluation_table}
\begin{tabular}{|c|c|c|c|c|}
\hline
Method          & Frame Rate & Input           & Success Ratio & RMSE[m] \\ \hline
ORB-SLAM3\cite{campos2021orb}  & 2Hz       & Front camera     & 0.000        & -    \\ \hline
ORB-SLAM3\cite{campos2021orb}  & 12Hz       & Front camera     & 0.787        & 0.199    \\ \hline
DROID-SLAM\cite{teed2021droid} & 2Hz        & Front camera     & 0.753        & 0.282    \\ \hline
\core(ours)                    & 2Hz        & Surround cameras &\textbf{0.993}        & \textbf{0.140}   \\ \hline\hline
\core(ours)                    & 2Hz        & 3D semantic occupancy ground truth*   & 1.000        & 0.122    \\ \hline
\end{tabular}
\begin{tablenotes}
    \footnotesize
    \item The best results are highlighted in the bold face. 
\end{tablenotes}
\end{threeparttable}
\end{table*}

%% file: table/ablation_study.tex
\begin{table*}[]
\centering
\begin{threeparttable}
\caption{Success Ratio and RMSE of APE [m] of \core for ablation study.}
\vspace{-1.0em}
\label{ablation_study_table}
\begin{tabular}{|c|c|c|c|c|c|}
\hline
Semantic Label Filter & Dynamic Object Filter & Label-based Object Filter & \pfilter & Success Ratio & RMSE[m] \\ \hline
             &              &             &             & 0.973 & 0.220   \\ \hline
\checkmark   &              &             &             & 0.973 & 0.206   \\ \hline
\checkmark   &              & \checkmark  &             & 0.973 & 0.218   \\ \hline
\checkmark   & \checkmark   &             &             & 0.980 & 0.173   \\ \hline
\checkmark   & \checkmark   &             & \checkmark  & \textbf{0.993} & \textbf{0.140}   \\ \hline
\end{tabular}
\begin{tablenotes}
    \footnotesize
    \item The best results are highlighted in the bold face. 
\end{tablenotes}
\end{threeparttable}
\vspace{-2.0em}
\end{table*}

%% file: table/map_compare.tex
\definecolor{barriercolor}{RGB}{255,0,0}
\definecolor{trafficconecolor}{RGB}{255,165,0}
\definecolor{drivablesurfacecolor}{RGB}{135,206,235}
\definecolor{otherflatcolor}{RGB}{165,42,42}
\definecolor{sidewalkcolor}{RGB}{211,211,211}
\definecolor{terraincolor}{RGB}{153,102,51}
\definecolor{manmadecolor}{RGB}{128,128,128}
\definecolor{vegetationcolor}{RGB}{34,139,34}

\newcolumntype{M}[1]{>{\centering\arraybackslash}p{#1}}

\begin{table*}[]
\centering
\caption{Partial visualization results on Occ3D-nuScenes validation sequences}
\vspace{-1.0em}
\label{map_compare_table}
\begin{tabular}{c M{0.55\columnwidth} M{0.55\columnwidth} M{0.55\columnwidth}}
Input Sequence & OCC-VO(ours) & OCC-VO(without \pfilter) & Ground Truth \\ 




Scene-0268     & 
\begin{minipage}[b]{0.55\columnwidth}\raisebox{-.45\height}{\includegraphics[width=\linewidth]{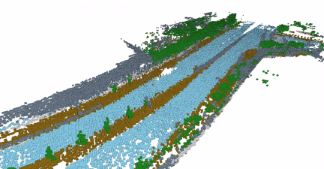}}\end{minipage} &
\begin{minipage}[b]{0.55\columnwidth}\raisebox{-.45\height}{\includegraphics[width=\linewidth]{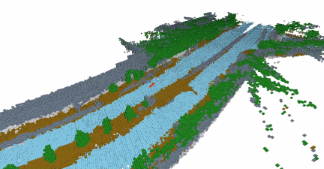}}\end{minipage} & 
\begin{minipage}[b]{0.55\columnwidth}\raisebox{-.45\height}{\includegraphics[width=\linewidth]{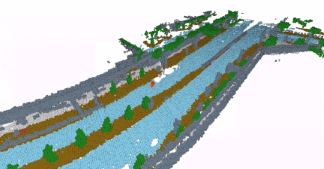}}\end{minipage} \\ 

Scene-0269     & 
\begin{minipage}[b]{0.55\columnwidth}\raisebox{-.45\height}{\includegraphics[width=\linewidth]{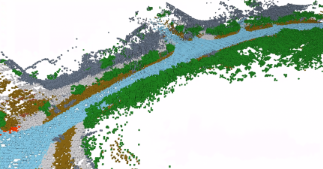}}\end{minipage} &
\begin{minipage}[b]{0.55\columnwidth}\raisebox{-.45\height}{\includegraphics[width=\linewidth]{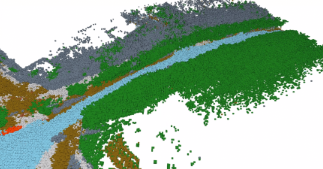}}\end{minipage} & 
\begin{minipage}[b]{0.55\columnwidth}\raisebox{-.45\height}{\includegraphics[width=\linewidth]{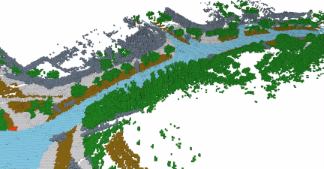}}\end{minipage} \\ 


Scene-0802     & 
\begin{minipage}[b]{0.55\columnwidth}\raisebox{-.45\height}{\includegraphics[width=\linewidth]{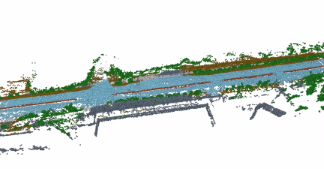}}\end{minipage} &
\begin{minipage}[b]{0.55\columnwidth}\raisebox{-.45\height}{\includegraphics[width=\linewidth]{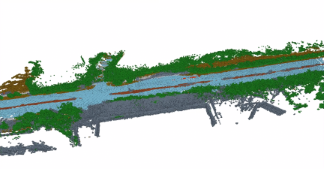}}\end{minipage} & 
\begin{minipage}[b]{0.55\columnwidth}\raisebox{-.45\height}{\includegraphics[width=\linewidth]{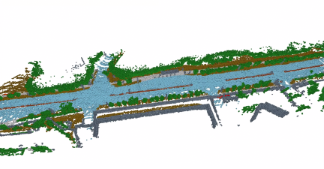}}\end{minipage} \\ 



\end{tabular}

\vspace{2pt}

\begin{tabular}{*{16}{@{}l}}
\raisebox{0.25ex}{\colorbox{barriercolor}{\rule{0pt}{1ex}\rule{1ex}{0pt}}} & \,Barrier \hspace{1em} &
\raisebox{0.25ex}{\colorbox{trafficconecolor}{\rule{0pt}{1ex}\rule{1ex}{0pt}}} & \,Traffic cone \hspace{1em} &
\raisebox{0.25ex}{\colorbox{drivablesurfacecolor}{\rule{0pt}{1ex}\rule{1ex}{0pt}}} & \,Driveable surface \hspace{1em} &
\raisebox{0.25ex}{\colorbox{otherflatcolor}{\rule{0pt}{1ex}\rule{1ex}{0pt}}} & \,Other flat \hspace{1em} &
\raisebox{0.25ex}{\colorbox{sidewalkcolor}{\rule{0pt}{1ex}\rule{1ex}{0pt}}} & \,Sidewalk \hspace{1em} &
\raisebox{0.25ex}{\colorbox{terraincolor}{\rule{0pt}{1ex}\rule{1ex}{0pt}}} & \,Terrain \hspace{1em} &
\raisebox{0.25ex}{\colorbox{manmadecolor}{\rule{0pt}{1ex}\rule{1ex}{0pt}}} & \,Manmade \hspace{1em} &
\raisebox{0.25ex}{\colorbox{vegetationcolor}{\rule{0pt}{1ex}\rule{1ex}{0pt}}} & \,Vegetation\\
\end{tabular}
\vspace{-2.0em}
\end{table*}

%% file: table/map_evaluation.tex
\begin{table}[]
\centering
\caption{The accuracy [m], precision [$<$0.4m] and completion ratio [$<$0.4m] of \core}
\vspace{-1.0em}
\label{map_evaluation_table}
\begin{tabular}{|c|c|c|c|}
\hline
Method                  & Acc.{[}m{]} & Precision & Comp. Ratio  \\ \hline
OCC-VO(ours)            & \textbf{0.111} & \textbf{0.724}      & 0.725       \\ \hline
OCC-VO(w/o \pfilter) & 0.125    & 0.572      & \textbf{0.875}      \\ \hline
\end{tabular}
\vspace{-2.5em}
\end{table}

%% file: table/runtime_analysis.tex
\begin{table}[]
\centering
\caption{Average execution time [ms/frame] of \core }
\vspace{-1.0em}
\label{execution_time_analysis_table}
\begin{tabular}{|cc|cc|}
\hline
\multicolumn{2}{|c|}{Module}                                                & \multicolumn{2}{c|}{Time{[}ms/frame{]}}               \\ \hline
\multicolumn{2}{|c|}{3D semantic occupancy prediction (A100)}                      & \multicolumn{2}{c|}{267}                        \\ \hline
\multicolumn{1}{|c|}{\multirow{5}{*}{Registration(i9-13900K)}} & GICP Algorithm        & \multicolumn{1}{c|}{69}  & \multirow{5}{*}{371} \\ \cline{2-3}
\multicolumn{1}{|c|}{}                              & Dynamic Object Filter & \multicolumn{1}{c|}{4}   &                      \\ \cline{2-3}
\multicolumn{1}{|c|}{}                              & GICP Algorithm        & \multicolumn{1}{c|}{70}  &                      \\ \cline{2-3}
\multicolumn{1}{|c|}{}                              & \pfilter               & \multicolumn{1}{c|}{166} &                      \\ \cline{2-3}
\multicolumn{1}{|c|}{}                              & Other                 & \multicolumn{1}{c|}{62}  &                      \\ \hline
\end{tabular}
\vspace{-2.5em}
\end{table}

%% file: chapters/conclusion.tex
\section{Conclusion} \label{Conclusion}

In our work, we introduce \core, a novel VO framework leveraging 3D semantic occupancy, making it distinct from traditional Visual SLAM. By using the designed filters, this innovation not only facilitates the generation of denser maps but also produces more accurate trajectories in autonomous driving scenarios. Our experiments on the Occ3D-nuScenes dataset demonstrate \core's superior performance in terms of accuracy and robustness in autonomous driving scenarios. In future work, we intend to integrate modules such as loop closure detection into \core, advancing it towards a SLAM system.